\documentclass[conference]{IEEEtran}

\usepackage[T1]{fontenc}

\usepackage[hyphens]{url}
\usepackage{graphicx}
\usepackage[numbers]{natbib}   
\usepackage{caption}

\usepackage{amsmath}
\usepackage{amssymb}
\usepackage{mathtools}
\usepackage{amsthm}
\usepackage{algorithm}
\usepackage{algorithmic}
\usepackage{booktabs}
\usepackage{multirow}
\usepackage{xcolor}
\usepackage{pifont}
\usepackage{subcaption}
\usepackage{enumitem}
\usepackage{listings}

\lstdefinestyle{compact}{basicstyle=\footnotesize\ttfamily}

\usepackage{hyperref}
\hypersetup{
    colorlinks=true,
    linkcolor=blue,
    filecolor=teal,      
    urlcolor=teal,
    citecolor=magenta,
    pdftitle={An Example},
    pdfpagemode=FullScreen,
    }

\definecolor{randcol}{RGB}{214,183,165}
\definecolor{kindonlycol}{RGB}{190,143,112}
\definecolor{kindsemcol}{RGB}{123,125,103}
\definecolor{fullfpcol}{RGB}{165,162,132}
\definecolor{theorycol}{RGB}{205,203,188}

\newenvironment{packeditemize}{
	\begin{list}{$\bullet$}{
			\setlength{\labelwidth}{4pt}
			\setlength{\itemsep}{0pt}
			\setlength{\leftmargin}{\labelwidth}
			\addtolength{\leftmargin}{\labelsep}
			\setlength{\parindent}{0pt}
			\setlength{\listparindent}{\parindent}
			\setlength{\parsep}{0pt}
			\setlength{\topsep}{1pt}}}{\end{list}}

\newcommand{\cmark}{\textcolor{green!60!black}{\ding{51}}}
\newcommand{\xmark}{\textcolor{red!70!black}{\ding{55}}}
\newcommand{\pmark}{\textcolor{orange!80!black}{$\circ$}}

\newcommand{\method}{Knows\xspace}
\usepackage{xspace}

\title{Knows: Agent-Native Structured Research Representations}

\author{
\IEEEauthorblockN{Guangsheng Yu}
\IEEEauthorblockA{Independent Researcher \\ knows.academy@gmail.com}
\and
\IEEEauthorblockN{Xu Wang}
\IEEEauthorblockA{Independent Researcher \\ knows.academy@gmail.com}
}

\begin{document}

\maketitle

\begin{abstract}
Research artifacts are distributed primarily as reader-oriented documents like PDFs. This creates a bottleneck for increasingly agent-assisted and agent-native research workflows, in which LLM agents need to infer fine-grained, task-relevant information from lengthy full documents, a process that is expensive, repetitive, and unstable at scale.

We introduce \method{}, a lightweight companion specification that binds structured claims, evidence, provenance, and verifiable relations to existing research artifacts in a form LLM agents can consume directly.
\method{} addresses the gap with a thin YAML sidecar (KnowsRecord) that coexists with the original PDF, requiring no changes to the publication itself, and validated by a deterministic schema linter.
We evaluate \method{} on 140 comprehension questions across 20 papers spanning 14 academic disciplines, comparing PDF-only, sidecar-only, and hybrid conditions across six LLM agents of varying capacity.
Weak models (0.8B--2B parameters) improve from 19--25\% to 47--67\% accuracy (+29 to +42 percentage points) when reading sidecar instead of PDF, while consuming 29--86\% fewer input tokens; an LLM-as-judge re-scoring confirms that weak-model sidecar accuracy (75--77\%) approaches stronger-model PDF accuracy (78--83\%). Beyond this controlled evaluation, a community sidecar hub at \url{https://knows.academy/} has already indexed over ten thousand publications and continues to grow daily, providing independent evidence that the format is adoption-ready at scale.
\end{abstract}

\section{Introduction}
Research progress is constrained not only by how quickly new knowledge is produced, but also by how efficiently existing knowledge can be consumed, organized, and reused. Both humans and Large Language Model (LLM) agents operate under bounded attention and memory, yet research artifacts are still primarily designed for human consumption. Research is published as narrative documents: typeset PDFs following standard templates, with structured sections, figures, tables, and bibliographic shorthands such as ``[12]'', serve human readers well, but do not expose their content in a form directly usable by LLM agents.

For LLM agents, accessing fine-grained information such as contributions, methods, experimental settings, or results typically requires recovering it from reader-oriented documents such as PDFs or HTML pages. This means extracting usable structure from the presentation layer: locating relevant passages, resolving references, and aggregating information across sections. The process is computationally expensive and token-intensive, especially when full documents must be repeatedly retrieved and reprocessed. Its quality is also unstable, because it depends on prompt formulation, retrieval conditions, and limited context windows, often yielding incomplete or inconsistent representations of the same source.

The problem becomes more pronounced with the rise of agent-assisted and increasingly agent-native research workflows. The same recovery process is repeated across agents and systems: different agents, or different downstream modules, read the same paper, parse the same document, and reconstruct the same underlying content independently. This duplicated effort introduces both inefficiency and inconsistency, turning literature interaction into a throughput bottleneck for large-scale analysis, collaborative research, and autonomous research pipelines. The assumption that LLM agents are a marginal audience is no longer safe: Stanford and BroadAI are co-hosting Agents4Science~\cite{agents4science2025}, the first open conference in which the authors, reviewers, and presenters are themselves AI agents. Human-oriented prose is no longer the only target format the scholarly record has to serve.

This mismatch mirrors an earlier challenge in web engineering. Before content negotiation, servers returned a single representation regardless of the client. The introduction of the HTTP \texttt{Accept} header allowed clients to request different representations of the same resource, such as HTML or JSON. No analogous mechanism exists for scholarly artifacts: humans and LLM agents are still served essentially the same document, despite fundamentally different consumption requirements.



We propose \method{}, a minimal companion sidecar specification for research artifacts. A KnowsRecord is a YAML file that coexists with the original PDF and binds author-asserted statements, quantitative evidence, typed relations, and provenance metadata in a schema-validated, agent-consumable format. The sidecar requires no changes to the published artifact; it composes with existing standards (DataCite for identifiers, PROV for provenance, CiTO for citation intent) rather than replacing them.

\begin{figure}[t]
  \centering
  \includegraphics[width=\columnwidth]{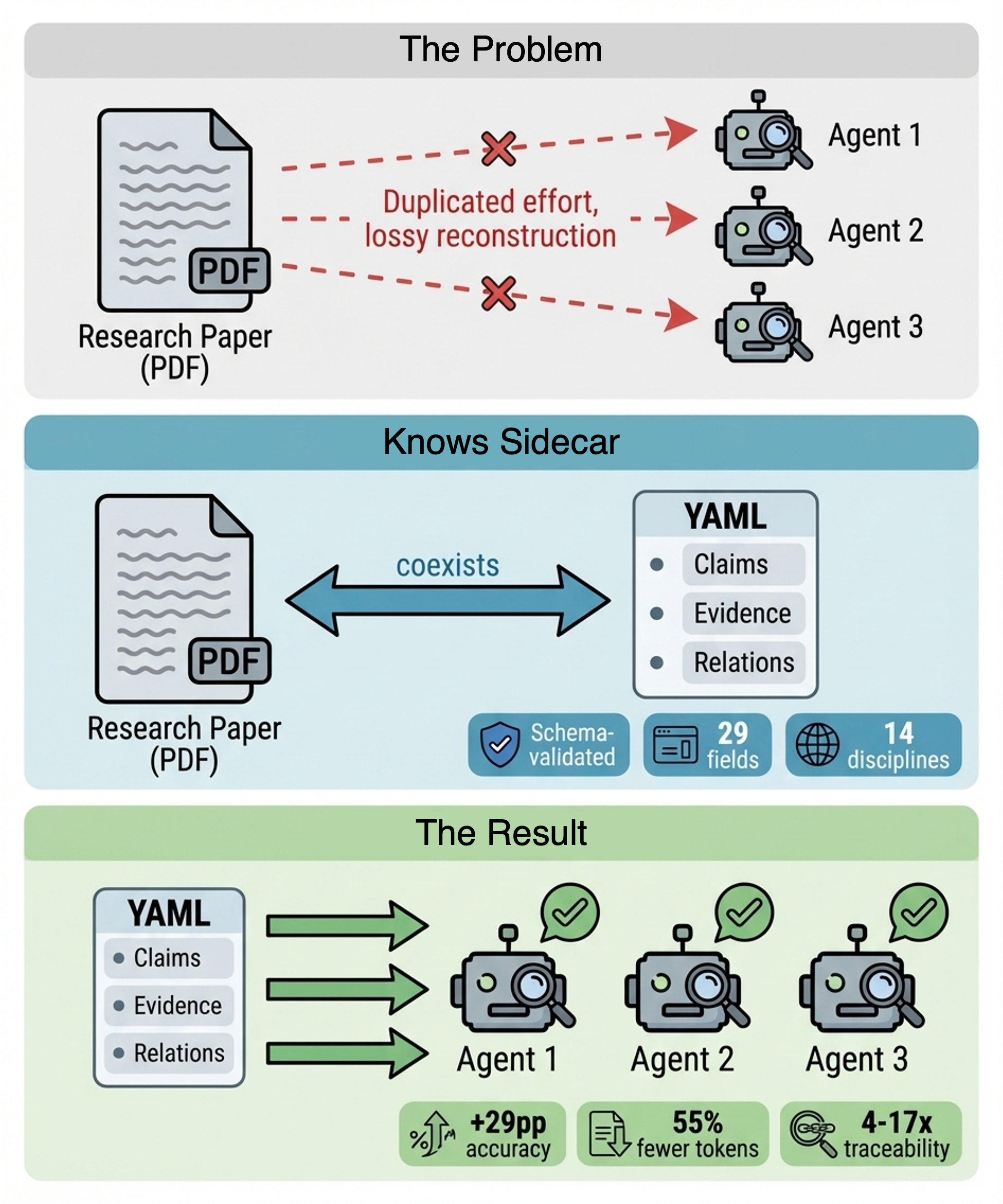}
  \caption{Overview of \method{}. A thin YAML sidecar (KnowsRecord) supplements the existing PDF with structured claims, evidence, and relations. Agents read the sidecar for structured queries (saving 29--86\% tokens) and fall back to the PDF only when sidecar coverage is insufficient. The same sidecar enables lint-based consistency validation, traceable reviews via cross-record references, and cross-record traceability through typed relation graphs.}
  \label{fig:overview}
\end{figure}

Our contributions are:
\begin{packeditemize}
\item The KnowsRecord specification (v0.9): 30 root-level fields, 23 entity definitions, tested across 14 academic disciplines from computer science to philosophy.
\item Reference tooling: a deterministic schema linter (\texttt{knows-lint}, seven validation checks), a sidecar scaffold generator (\texttt{knows-gen}) that ingests either the author's manuscript source or a rendered PDF, and a unified CLI with six subcommands.
\item A comprehensive evaluation on 140 questions across 20 papers with six LLM agents in three capacity tiers, demonstrating that sidecar-assisted agents match PDF accuracy with 55\% fewer tokens and that weak models gain +29 to +42 percentage points.
\item A review-as-sidecar workflow, in which structured reviews are themselves KnowsRecords linked to the original paper via cross-record relations and yield 4--17$\times$ more traceable reasoning.
\item A deployed community sidecar hub at \url{https://knows.academy/} that has already indexed over ten thousand publications and continues to grow daily, providing real-world deployment evidence that complements the controlled evaluation in this report.
\end{packeditemize}

\section{Background and Related Work}

\subsection{Machine-Readable Scholarly Metadata}

Structured metadata for research has a long history. DataCite provides persistent identifiers and basic bibliographic fields. The FAIR principles formalize expectations for findability, accessibility, interoperability, and reusability of research data. Nanopublications decompose scientific assertions into RDF triples with provenance that form a formal claim graph, but their adoption has remained limited to specialized communities because RDF authoring is costly~\cite{groth2010nanopub}.

The Open Research Knowledge Graph (ORKG) organizes research contributions as structured comparison tables, focusing on born-digital data rather than retrospective annotation of existing papers \cite{jaradeh2019orkg}. Schema.org and its scholarly extensions provide vocabulary for web-scale metadata but operate at the bibliographic level, not at the granularity of individual claims and evidence.

\subsection{LLM Agents and Paper Consumption}

The emergence of LLM-based research agents has created new demand for machine-stable scholarly metadata. Paper2Agent converts papers into MCP tool endpoints post hoc, so an agent can interact with individual papers through a retrieval pipeline \cite{miao2025paper2agent}. Agentic Publication (AP) wraps papers in agent-accessible interfaces but does not define a reusable schema for the underlying metadata \cite{pugliese2025agenticpub}. AgentRxiv provides a distribution platform for agent-consumable preprints without specifying a companion metadata format \cite{schmidgall2025agentrxiv}.

These approaches share a common limitation: they reconstruct structure from prose at consumption time, incurring extraction loss and duplicated effort across consumers.

\subsection{Autonomous Research Agents}
\label{sec:autonomous-research}

A growing class of autonomous systems now authors, reviews, or explores scientific ideas with minimal human intervention. The AI Scientist~\cite{lu2024aiscientist} orchestrates end-to-end paper generation including literature search, experimentation, and write-up. NovelSeek~\cite{novelseek2025} and Dolphin~\cite{dolphin2025} extend this pattern with closed-loop refinement of hypotheses against empirical results. CodeScientist~\cite{codescientist} emphasizes code-based experimentation, Data-to-Paper~\cite{datatopaper} targets reproducible data analysis, and EvoScientist~\cite{evoscientist} applies evolutionary search over research directions. A broader survey of AI-generated science is catalogued in~\cite{babyaigs2024}, and recent benchmarks such as ResearchClawBench~\cite{researchclawbench} evaluate autonomous research agents across multi-stage workflows. At the level of capability organization, these systems are increasingly described in terms of \emph{agentic skills}, namely reusable and composable action repertoires that extend beyond single tool calls~\cite{jiang2026sokagenticskills}; the structured scholarly record is the natural interchange format at that level.

These systems share one property: they already \emph{author} and \emph{consume} research in structured internal representations (typed claims, experiment plans, result tables, citation graphs), and the PDF is only produced as a terminal projection for human readers. When two such systems interact (one producing a paper, another reviewing or building on it), prose becomes a lossy serialization in the middle. The specification described in this report is what a shared agent-native metadata layer between such systems looks like.

\subsection{Positioning}

\method{} fills a specific gap between bibliographic metadata (DataCite, Schema.org) and full-text extraction pipelines (Paper2Agent, AP). Table~\ref{tab:comparison} compares \method{} against existing approaches on six dimensions that distinguish agent-native metadata from general scholarly infrastructure, and Table~\ref{tab:compat} summarizes the integration stance toward widely deployed scholarly metadata standards.

\begin{table}[t]
\centering
\caption{Qualitative comparison of \method{} with existing approaches. \cmark{} = full support, \pmark{} = partial, \xmark{} = absent.}
\label{tab:comparison}
\resizebox{\columnwidth}{!}{%
\begin{tabular}{lcccccc}
\toprule
\textbf{System} & \textbf{Schema} & \textbf{Author} & \textbf{Claim} & \textbf{Cross-} & \textbf{Action} & \textbf{Lint} \\
& \textbf{Valid.} & \textbf{Time} & \textbf{DAG} & \textbf{Paper} & \textbf{Hooks} & \textbf{Check} \\
\midrule
Nanopub & \cmark & \pmark & \cmark & \cmark & \xmark & \xmark \\
ORKG & \cmark & \pmark & \pmark & \cmark & \xmark & \xmark \\
Paper2Agent & \xmark & \xmark & \xmark & \xmark & \cmark & \xmark \\
AP & \xmark & \xmark & \xmark & \xmark & \xmark & \xmark \\
AgentRxiv & \xmark & N/A & \xmark & \xmark & \xmark & \xmark \\
\textbf{\method{}} & \cmark & \cmark & \cmark & \cmark & \cmark & \cmark \\
\bottomrule
\end{tabular}%
}
\end{table}

\begin{table*}[t]
\centering
\small
\caption{Integration stance of \method{} toward widely deployed scholarly metadata standards. \method{} does not replace any of these; it \emph{points to} identifier and artefact metadata via \texttt{identifiers} and \texttt{external\_metadata\_refs}, and adds a claim-, evidence-, and relation-layer that none of them expose.}
\label{tab:compat}
\resizebox{\linewidth}{!}{%

\begin{tabular}{lllll}
\toprule
\textbf{Standard} & \textbf{Layer} & \textbf{Role in \method{}} & \textbf{Mechanism} & \textbf{v0.9 status} \\
\midrule
DOI & identifier & consume & \texttt{identifiers.doi} & \cmark{} \\
arXiv & identifier & consume & \texttt{identifiers.arxiv} & \cmark{} \\
ISBN & identifier & consume & \texttt{identifiers.isbn} & \cmark{} \\
ORCID & identifier & consume & \texttt{authors[].orcid} & \cmark{} \\
\midrule
DataCite & bibliographic & reference & \texttt{external\_metadata\_refs.standard=datacite} & \cmark{} \\
Schema.org & bibliographic & compatible (not consumed) & out-of-band JSON-LD & \pmark{} \\
Dublin Core / FAIR & bibliographic & principles-aligned & no direct field mapping & \pmark{} \\
\midrule
CodeMeta & artefact (repo) & reference & \texttt{external\_metadata\_refs.standard=codemeta} & \cmark{} \\
CITATION.cff & artefact (code) & reference & \texttt{external\_metadata\_refs.standard=citation\_cff} & \cmark{} \\
Croissant & artefact (dataset) & reference & \texttt{external\_metadata\_refs.standard=croissant} & \cmark{} \\
Model Card & artefact (model) & reference & \texttt{external\_metadata\_refs.standard=model\_card} & \cmark{} \\
RO-Crate & artefact (bundle) & reference & \texttt{external\_metadata\_refs.standard=ro\_crate} & \cmark{} \\
\midrule
BibTeX & citation format & convertible & importer/exporter deferred to tooling & \pmark{} \\
CSL-JSON & citation format & convertible & importer/exporter deferred to tooling & \pmark{} \\
Nanopub & semantic claim & conceptually aligned & statement+relation graph analogue & \pmark{} \\
PROV & provenance & aligned & \texttt{provenance} record with origin/actor/method & \pmark{} \\
CiTO & citation intent & aligned & \texttt{Relation.citation\_intent} enum & \pmark{} \\
\bottomrule
\end{tabular}%
}
\end{table*}

The integration stance is deliberately conservative: \method{} consumes identifier schemes directly, references artefact-level metadata standards via a typed pointer (\texttt{external\_metadata\_refs}), and leaves BibTeX/CSL-JSON round-trips to tooling rather than absorbing them into the core schema. The claim-, evidence-, and relation-level structure that \method{} adds has no direct counterpart in any of the referenced standards, which motivates the specification rather than a reuse of existing vocabularies.

\section{The KnowsRecord Specification}

A KnowsRecord is a YAML document that conforms to a JSON Schema (v0.9, 30 root-level fields, 23 entity definitions). It coexists with the original artifact as a companion file (e.g., \texttt{paper.knows.yaml} alongside \texttt{paper.pdf}).

\smallskip\noindent\textbf{Core Entities.}
The record contains five primary collections. \emph{Artifacts} declare the paper and its cited works with typed identifiers. \emph{Statements} capture claims, assumptions, limitations, methods, questions, and definitions, each annotated with a modality (empirical, theoretical, descriptive, normative) and a two-dimensional confidence score: \emph{claim\_strength} (how strongly the author asserts the claim) and \emph{extraction\_fidelity} (how faithfully the metadata represents the source). \emph{Evidence} items link to specific observations with numeric values or qualitative descriptions. \emph{Relations} form a typed directed graph connecting statements, evidence, and artifacts via predicates such as \texttt{supported\_by}, \texttt{challenged\_by}, \texttt{depends\_on}, and \texttt{cites}. \emph{Actions} define optional executable hooks with mandatory safety policies.

Table~\ref{tab:entities} summarizes the required fields and the key enumerations for the five primary entities; full per-entity schemas appear in Appendix~\ref{app:schema} and in the accompanying open-source repository.

\begin{table*}[t]
\centering
\resizebox{\linewidth}{!}{%
\begin{tabular}{@{}lp{0.32\textwidth}p{0.48\textwidth}@{}}
\toprule
\textbf{Entity} & \textbf{Required fields} & \textbf{Key enumerations} \\
\midrule
\emph{Artifact}  & \texttt{id} (prefix \texttt{art:}), \texttt{artifact\_type}, \texttt{role}, \texttt{title} &
\texttt{artifact\_type} $\in$ \{paper, repository, dataset, model, benchmark, software, website, other\}; \texttt{role} $\in$ \{subject, supporting, cited\} \\
\emph{Statement} & \texttt{id} (prefix \texttt{stmt:}), \texttt{statement\_type}, \texttt{text}, \texttt{about\_ref}, \texttt{status} &
\texttt{statement\_type} $\in$ \{claim, assumption, limitation, method, question, definition\}; \texttt{status} $\in$ \{asserted, retracted, superseded, under\_review\}; optional \texttt{modality} $\in$ \{empirical, theoretical, descriptive, normative\} \\
\emph{Evidence}  & \texttt{id} (prefix \texttt{ev:}), \texttt{evidence\_type}, \texttt{summary} &
\texttt{evidence\_type} $\in$ \{table\_result, figure, experiment\_run, citation\_backed, artifact\_run, proof, case\_study, clinical\_trial, survey\_result, qualitative\_analysis, statistical\_test, observation, simulation, other\} \\
\emph{Relation}  & \texttt{id} (prefix \texttt{rel:}), \texttt{subject\_ref}, \texttt{predicate}, \texttt{object\_ref} &
\texttt{predicate} $\in$ \{supported\_by, challenged\_by, depends\_on, limited\_by, uses, evaluates\_on, implements, documents, cites, same\_as, supersedes, retracts\}; optional \texttt{citation\_intent} $\in$ \{supports, extends, uses\_method, compares\_to, contradicts, reviews, cites\_data, background, other\} \\
\emph{Action}    & \texttt{id} (prefix \texttt{act:}), \texttt{action\_type}, \texttt{label}, \texttt{target\_ref}, \texttt{interface}, \texttt{safety} &
\texttt{action\_type} $\in$ \{download, run, query, deploy, other\}; \texttt{safety.side\_effects} $\in$ \{none, temporary\_files\_only, persistent, external\}; when \texttt{safety.sandbox\_required}=true, consumers MUST execute in isolation \\
\bottomrule
\end{tabular}
}
\caption{Required fields and key enumerations for the five primary entities in a KnowsRecord. Every entity carries the same ID-prefix discipline so that cross-references resolve syntactically; lint rejects any reference whose prefix does not match the target entity type.}
\label{tab:entities}
\end{table*}

\smallskip\noindent\textbf{Provenance and Versioning.}
Each record carries provenance metadata (origin, actor, method, verification status) and a version triple separating the specification version, the record version, and the source version. An optional \texttt{replaces} field points to the \texttt{record\_id} of the previous version, forming a singly-linked version chain that agents traverse to access revision history; the old record sets \texttt{record\_status} to \texttt{superseded}. A freshness block declares the record's temporal validity with \texttt{as\_of}, \texttt{update\_policy}, and optional \texttt{stale\_after} fields.

\smallskip\noindent\textbf{Cross-Record References.}
Entities in different KnowsRecords reference each other via the grammar \texttt{record\_id\#local\_id}. This enables cross-record traceability: a review sidecar (\texttt{profile: review@1}) can link its weakness statements directly to specific claims in the reviewed paper's sidecar. Unstructured reasoning across paper bodies (e.g., joint comprehension of two papers' narratives) remains a consumer-side task and is out of scope for the specification.

\smallskip\noindent\textbf{Extensibility.}
Twenty of 23 entity definitions include an \texttt{x\_extensions} field for forward-compatible customization without breaking schema validation. The root object enforces \texttt{additionalProperties: false} for structural integrity, while entities remain extensible.

\smallskip\noindent\textbf{Profile Namespaces and Backward Compatibility.}
The specification follows semantic versioning: a major bump indicates a breaking change to required fields or enumerations; a minor bump adds optional fields or predicates without invalidating existing records; a patch bump is editorial. Each record declares its target spec version in \texttt{knows\_version}, and readers MUST refuse records whose major version exceeds the reader's. Domain extensions live in the \texttt{profile} field (e.g., \texttt{paper@1}, \texttt{review@1}, \texttt{dataset@1}), which carries its own major version independent of the core spec; adding a new profile is never a breaking change. A reader that does not recognize a profile MAY still validate the core record and expose profile-specific data as opaque \texttt{extensions}. For forward compatibility within a major version, readers MUST NOT reject records containing unknown \texttt{x\_extensions} keys and MUST preserve them when re-emitting.

\smallskip\noindent\textbf{Trust Model and Threat Assumptions.}
A KnowsRecord is an author assertion, not a certification. We distinguish three actors: the \emph{author}, who asserts statements, evidence, and provenance; the \emph{consumer}, typically an LLM agent that reads the record; and an optional \emph{third-party verifier} such as a reviewer, a sidecar hub, or a downstream attestor. Consumers MAY treat structural properties (field presence, ID uniqueness, cross-reference resolution, and schema conformance) as verifiable by the \texttt{knows-lint} tool alone, and SHOULD refuse records that fail structural verification. Semantic properties, including whether a reported accuracy actually appears in the PDF or whether a cited work supports the intended claim, are not verified by the specification and remain the authoring party's responsibility. We distinguish two classes of adversarial behaviour. \emph{Structural corruption} (malformed records, dangling references, invented field names) is caught by the linter and fails fast. \emph{Semantic corruption} (inflated confidence, fabricated observations, misattributed citations) is out of scope for v0.9, and the specification makes no normative claim that consumers will detect it; the reference implementation treats a structurally well-formed record as authoritative at the semantic level. The broader question of how much a consumer should trust artefacts produced by an open agentic pipeline is analysed in~\cite{chen2026clawed}, and the mitigations applicable to Knows records are consistent with the defensive posture recommended there. Specific countermeasures are discussed in \S\ref{sec:discussion} and include LLM-based cross-checking of evidence against source anchors, and signed attestation for records promoted to a shared hub. Cryptographic signing, revocation, and public-key infrastructure are explicitly out of scope: a signed record can be encoded under \texttt{x\_extensions} on \texttt{provenance}, but the specification does not mandate a scheme.

\smallskip\noindent\textbf{Governance and Evolution.}
The KnowsRecord specification is maintained by a single canonical maintainer group, coordinated through the project hub at \url{https://knows.academy/}, and released under a CC-BY-4.0 licence that is distinct from the licence of each individual sidecar record (every record declares its own \texttt{license} field). Breaking changes to required fields or enumerations require a major version bump and follow an RFC process announced through the project hub; additive changes such as new optional fields, new predicate enum values, or new profiles ship in minor releases without requiring reader upgrades. Domain profiles under \texttt{profile: <name>@<major>} may be proposed and maintained independently of the core specification; a cross-profile registry MAY be added in a future revision but is out of scope for v0.9. The reference implementation, schema, canonical consumption prompt, and experiment harness are released under the same repository, and this technical report is itself distributed together with its own KnowsRecord sidecar.

\begin{figure*}[t]
  \centering
  \includegraphics[width=\textwidth]{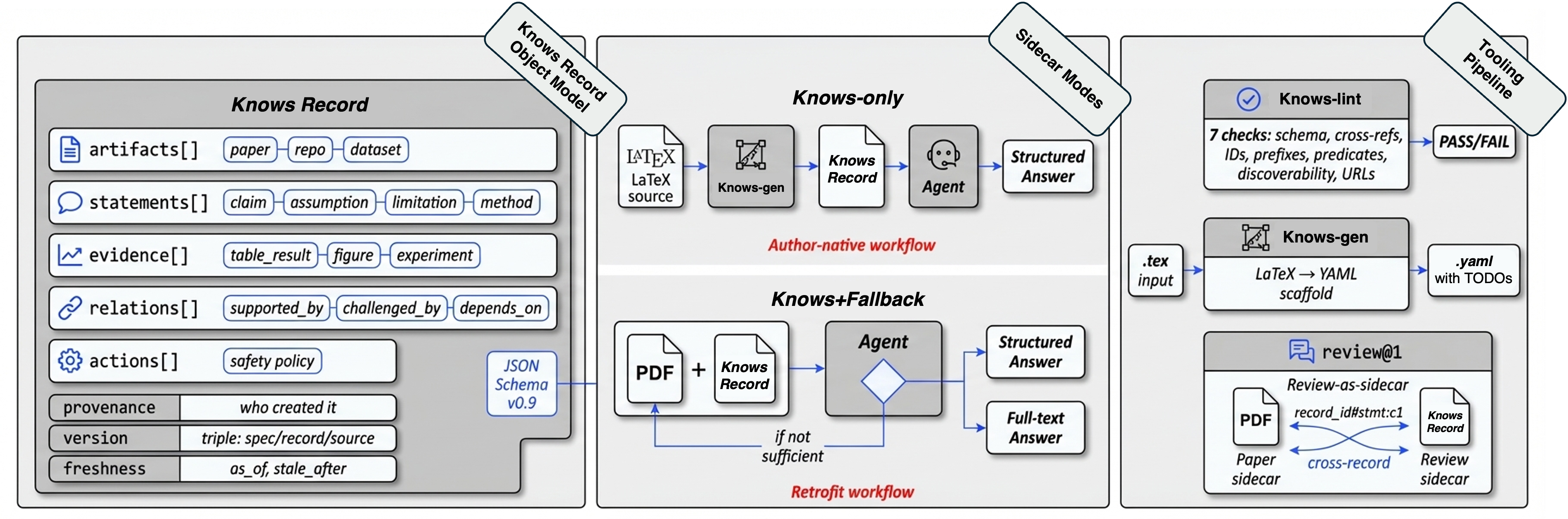}
  \caption{Architecture of the KnowsRecord specification and tooling ecosystem. Left: the KnowsRecord object model with five primary entity collections (artifacts, statements, evidence, relations, actions) plus metadata layers (provenance, version, freshness). Center: the two sidecar modes, author-native (Knows-only) and retrofit (Knows+Fallback). Right: the tooling pipeline including \texttt{knows-gen} for scaffold generation from authoring sources or PDFs, \texttt{knows-lint} for seven-check schema validation, and the review-as-sidecar workflow with cross-record relation linking.}
  \label{fig:architecture}
\end{figure*}

\section{Tooling and Workflow}

\smallskip\noindent\textbf{\texttt{knows-lint}.}
The linter performs seven deterministic checks: JSON Schema validation, cross-reference integrity, ID uniqueness, ID prefix conventions (\texttt{art:}, \texttt{stmt:}, \texttt{ev:}, \texttt{rel:}), relation predicate constraints, artifact discoverability, and optional URL liveness. All structural corruptions (missing fields, broken cross-references, removed entities) are caught; semantic corruptions (wrong numeric values, inflated confidence levels) are not, motivating future LLM-based semantic verification (\S\ref{sec:discussion}).

\smallskip\noindent\textbf{\texttt{knows-gen}.}
The scaffold generator operates on either of two inputs: an authoring source bundle (for example, a multi-file manuscript with nested include directives, multi-format author blocks, and extended citation commands) or a rendered PDF. In the source-bundle path the generator reconstructs the document's logical structure directly; in the PDF path it extracts claims, evidence, and citation targets via an LLM pass. Both paths produce a KnowsRecord with template placeholders that authors or retrofit agents complete, reducing hand-authoring effort to approximately 15 minutes for a typical conference paper. The experiments reported in \S\ref{sec:experiments} use the PDF path because most benchmark papers have no recoverable source.

\smallskip\noindent\textbf{Two Operating Modes.}
We define two sidecar usage modes. In the \emph{Knows-only} mode, the agent reads only the sidecar, which provides structured context at a fraction of the PDF token cost. In the \emph{Knows+Fallback} mode, the agent reads the sidecar first and falls back to the full PDF when the initial answer is insufficient. The fallback mode suits retrofit scenarios where sidecar coverage may be incomplete.

\smallskip\noindent\textbf{Agent Consumption Protocol.}
The two modes are configurations of a single retrieval protocol, shown in Algorithm~\ref{alg:consume}. Given a query $q$ and an artifact $a$, the consumer first attempts to load the companion sidecar $S$. If $S$ passes structural validation (\texttt{knows-lint}), the consumer ranks $S.\text{statements} \cup S.\text{evidence}$ by relevance to $q$ and resolves the referenced \texttt{source\_anchors} to obtain candidate localizations in the PDF. In \emph{Knows-only} mode the consumer returns the composed answer together with its sidecar-grounded citation trace. In \emph{Knows+Fallback} mode the consumer additionally consults the PDF when the sidecar-only answer falls below a confidence threshold~$\tau$, using the resolved anchors to localize retrieval to the relevant page, section, or figure. When no sidecar is available or structural validation fails, the protocol reduces to the baseline PDF read.

\begin{algorithm}[t]
\caption{Agent Consumption Protocol}
\label{alg:consume}
\begin{algorithmic}[1]
\REQUIRE query $q$; artifact identifier $a$; mode $m \in \{\textsc{only}, \textsc{fallback}\}$; threshold $\tau$
\ENSURE answer $A$ with citation trace $T$
\STATE $S \leftarrow \textsc{FetchSidecar}(a)$
\IF{$S = \bot$ \OR $\neg\,\textsc{Lint}(S)$}
    \RETURN $\textsc{BaselinePdf}(a, q)$
\ENDIF
\STATE $R \leftarrow \textsc{RetrieveTopK}(S.\text{statements} \cup S.\text{evidence},\; q)$
\STATE $\mathcal{A} \leftarrow \bigcup_{r \in R} r.\text{source\_anchors}$
\STATE $(A_0, c_0) \leftarrow \textsc{Compose}(R, \mathcal{A})$
\IF{$m = \textsc{only}$ \OR $c_0 \geq \tau$}
    \RETURN $(A_0,\; \textsc{Trace}(R, \mathcal{A}))$
\ENDIF
\STATE $P \leftarrow \textsc{FetchPassages}(a, \mathcal{A})$
\STATE $(A_1, c_1) \leftarrow \textsc{Compose}(R, \mathcal{A}, P)$
\RETURN $(A_1,\; \textsc{Trace}(R, \mathcal{A}, P))$
\end{algorithmic}
\end{algorithm}

\begin{figure}[t]
  \centering
  \includegraphics[width=\columnwidth]{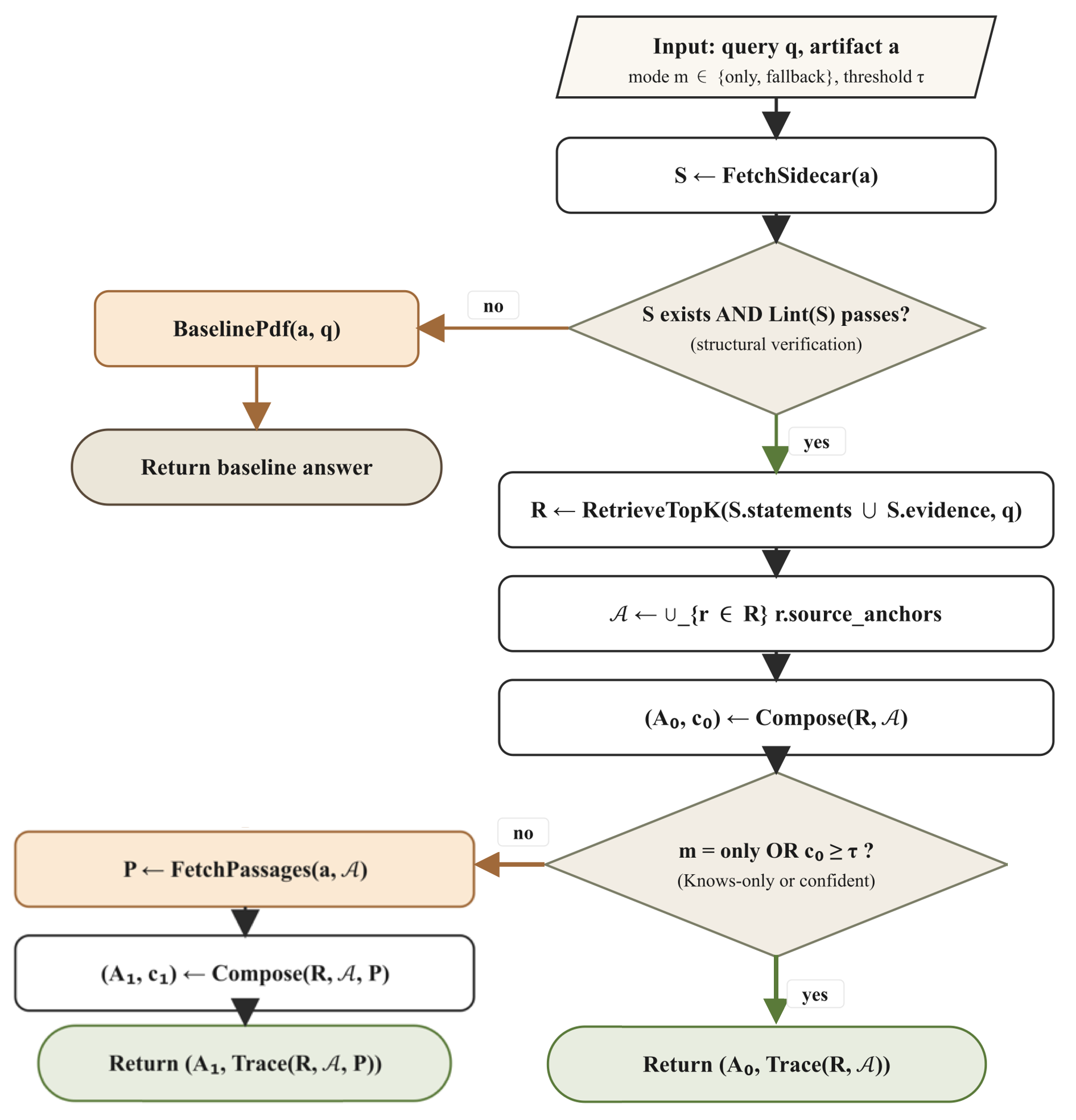}
  \caption{Agent Retrieval Flow. Rounded rectangles are process steps, diamonds are decisions, and oval terminals denote return points. Warm-peach nodes (\texttt{BaselinePdf}, \texttt{FetchPassages}) mark the two cases where the consumer escapes the sidecar-only path: structural validation failure routes to the baseline PDF read, and sub-threshold confidence in \emph{Knows+Fallback} mode triggers PDF passage retrieval anchored by $\mathcal{A}$. Olive-green terminals are sidecar-grounded returns; the legend colors match the branch semantics. This figure visualizes the same normative protocol expressed by Algorithm~\ref{alg:consume}.}
  \label{fig:flow}
\end{figure}

\smallskip\noindent\textbf{Protocol vs Reference Implementation.}
Algorithm~\ref{alg:consume} specifies the normative consumption protocol, including the confidence threshold $\tau$ that governs when \emph{Knows+Fallback} mode consults the PDF. The canonical LLM prompt that realizes \textsc{Compose} is released as part of the Knows skill in the open-source repository; both the sidecar and the PDF conditions derive their prompts from the same document, so the only legitimate asymmetry between conditions is the content of \texttt{\{context\}}. The reference implementation evaluated in \S\ref{sec:experiments} substitutes $\tau$ with a hedging-phrase heuristic: fallback is triggered when the sidecar-only answer contains explicit signals of uncertainty (``insufficient evidence'', ``unclear from the sidecar''). We retain Algorithm~\ref{alg:consume} as the specification because a calibrated confidence signal is the correct long-run interface; the heuristic is a practical stand-in pending a cross-model confidence-probe study, which we defer to future work. The \texttt{knows query} CLI shipped with the reference toolchain is an offline keyword-match convenience and is not used in any evaluation; all experiments in \S\ref{sec:experiments} invoke the LLM consumption prompt directly.

\smallskip\noindent\textbf{Retrieval Order, Caching, and Conflict Resolution.}
The protocol prescribes a layered retrieval order: coarse-grained \emph{statements} supply the semantic skeleton, \emph{evidence} entries attach quantitative measurements, and \emph{source\_anchors} expose page-, section-, or figure-level pointers for fallback. Consumers SHOULD cache sidecars keyed by \texttt{record\_id} and \texttt{version.record}, and SHOULD invalidate cached entries once the current time passes the record's \texttt{freshness.stale\_after}. When a sidecar's quantitative claim disagrees with the PDF, the PDF is authoritative; when structural information such as a typed citation intent or a cross-record relation has no unambiguous counterpart in the PDF, the sidecar is authoritative, since structural assertions are explicitly authored rather than inferred from prose.

\smallskip\noindent\textbf{Review-as-Sidecar.}
Reviews are KnowsRecords with \texttt{profile: review@1}. Each review weakness links to specific original claims via cross-record relations (\texttt{record\_id\#stmt:c1}), so peer review becomes machine-traversable and per-weakness traceable.

\section{Experiments}
\label{sec:experiments}

We evaluate \method{} across ten experiments (E1--E10) covering accuracy, token efficiency, traceability, consistency validation, cross-record traceability, generation quality and generator--consumer cross-evaluation, coverage--accuracy ablation, statement granularity, and matched-output protocol validation.

\subsection{Experimental Settings}

\smallskip\noindent\textbf{Benchmark.}
We construct a benchmark of 140 L1 comprehension questions across 20 classic papers spanning 14 disciplines (computer science, biology, chemistry, physics, economics, psychology, medicine, education, philosophy, mathematics, civil engineering, mechanical engineering, electrical engineering, semiconductor engineering). Papers range from 579 words (Bloom 1956) to 198K words (Terzaghi 1943) and are categorized into four length groups: SHORT ($<$2K words, 3 papers), MEDIUM (2--8K, 6 papers), STANDARD (8--20K, 7 papers), and LONG ($>$20K, 4 papers). Each question has a ground-truth answer and a \texttt{sidecar\_answerable} tag indicating whether the sidecar contains sufficient information.

\smallskip\noindent\textbf{Models.}
We use six LLM agents across three capacity tiers: \emph{Weak} (Qwen3.5-0.8B, Qwen3.5-2B, local GPU), \emph{Medium} (MiMO-V2-Flash, Qwen3.5-27B, via OpenRouter), and \emph{Strong} (MiMO-V2-Pro, Kimi-K2.5, via OpenRouter).

\smallskip\noindent\textbf{Conditions.}
Three conditions: PDF-only (full paper text), Knows-only (sidecar YAML), and Knows+Fallback (sidecar first, PDF if insufficient). Scoring uses keyword overlap with 40\% phrase-match threshold, cross-validated by an LLM-as-judge (Claude Sonnet 4) on the full 1,200-answer corpus.

\smallskip\noindent\textbf{Data and Code Availability.}
The reference implementation, schema, canonical consumption prompt, benchmark questions, sidecar records, and evaluation harness are released at \url{https://knows.academy/}. Per-question responses, parsed outputs, token and latency logs, and random seeds are committed with the repository. The matched-output PDF rerun specification (the full-scale successor to the E10 pilot), the audit-trace validity rubric, the coverage-conditioned analysis protocol, and the release rules that govern how experimental outcomes drive the final narrative are locked in a pre-registered release-rules document distributed alongside this report. Sidecars were generated using the \texttt{knows-sidecar} package (v0.9), with the v1.0 base consumption prompt used for E1--E9 and the v1.1 matched-output prompt used for E10.

\subsection{Results}

\subsubsection{E1: Task Accuracy by Paper Length}

Table~\ref{tab:e1_length} reports accuracy stratified by paper length, the primary analysis dimension. Paper length is the strongest predictor of \method{} advantage for weak models.

\begin{table}[t]
\centering
\caption{E1 accuracy (\%) by paper length and model tier. $\Delta$ = Knows $-$ PDF. Weak models improve across all lengths; the advantage scales with paper length.}
\label{tab:e1_length}
\resizebox{\columnwidth}{!}{%
\begin{tabular}{lcrrrrrr}
\toprule
\textbf{Length} & \textbf{\#P/\#Q} & \multicolumn{2}{c}{\textbf{Weak}} & \multicolumn{2}{c}{\textbf{Medium}} & \multicolumn{2}{c}{\textbf{Strong}} \\
\cmidrule(lr){3-4} \cmidrule(lr){5-6} \cmidrule(lr){7-8}
& & PDF & $\Delta$ & PDF & $\Delta$ & PDF & $\Delta$ \\
\midrule
SHORT & 3/18 & 58 & +8 & 50 & +8 & 61 & +14 \\
MEDIUM & 6/41 & 24 & +29 & 73 & $-$6 & 66 & $-$5 \\
STANDARD & 7/58 & 15 & +40 & 81 & $-$19 & 67 & $-$9 \\
LONG & 4/23 & 7 & +57 & 48 & +13 & 52 & +17 \\
\midrule
\textbf{Overall} & 20/140 & 22 & +35 & 69 & $-$6 & 64 & $-$1 \\
\bottomrule
\end{tabular}%
}
\end{table}

\begin{center}
\fbox{\begin{minipage}{0.9\columnwidth}
\textbf{Takeaway (length effect).} Weak models improve at every paper length (+8 to +57 pp), with the largest gains on long papers where PDF context overwhelms small context windows. Medium models favor PDF on STANDARD papers ($-$19 pp) where fine-grained details are lost in the sidecar. Strong models show near-parity across lengths.
\end{minipage}}
\end{center}

Table~\ref{tab:discipline} controls for paper length (2K--20K words only) and compares accuracy across 10 disciplines, broken down by model tier. Weak models improve across all ten disciplines (+10 to +70 pp). Medium and strong models show discipline-dependent patterns: physics, chemistry, semiconductor, and biology favor Knows for strong models, while psychology, medicine, and CS favor PDF.

\begin{table}[t]
\centering
\caption{Cross-discipline accuracy $\Delta$ (Knows $-$ PDF, pp) on length-controlled papers (2K--20K words, 13 papers, 10 disciplines), by model tier.}
\label{tab:discipline}
\resizebox{\columnwidth}{!}{%
\begin{tabular}{lrrrr}
\toprule
\textbf{Discipline} & \textbf{\#Q} & \textbf{Weak $\Delta$$\uparrow$} & \textbf{Med. $\Delta$} & \textbf{Str. $\Delta$} \\
\midrule
Physics & 5 & \textbf{+70} & $-$20 & +10 \\
Chemistry & 13 & \textbf{+38} & $-$8 & \textbf{+15} \\
Semiconductor & 13 & \textbf{+27} & +4 & \textbf{+15} \\
Economics & 8 & \textbf{+38} & \textbf{+19} & $-$19 \\
Biology & 8 & \textbf{+38} & $-$6 & +6 \\
CS & 26 & \textbf{+40} & $-$25 & $-$12 \\
Mathematics & 5 & \textbf{+40} & $-$30 & $-$20 \\
Philosophy & 8 & \textbf{+31} & $-$12 & $-$31 \\
Psychology & 8 & \textbf{+19} & $-$31 & $-$31 \\
Medicine & 5 & \textbf{+10} & $-$30 & $-$40 \\
\midrule
\textbf{Overall} & 99 & \textbf{+35} & $-$14 & $-$8 \\
\bottomrule
\end{tabular}%
}
\end{table}

\subsubsection{E1 (continued): Per-Model Results and Scoring Validation}

Table~\ref{tab:e1_overall} shows per-model results with keyword scoring, LLM-judge validation, token savings, and fallback accuracy.

\begin{table}[t]
\centering
\caption{E1 per-model accuracy (\%). KW = keyword; LLM = Claude Sonnet 4 judge (correct + partial); FB = Knows+Fallback (15 original papers). Latency: average per-question response time.}
\label{tab:e1_overall}
\resizebox{\columnwidth}{!}{%
\begin{tabular}{llrrrrrrr}
\toprule
\textbf{Model} & \textbf{Tier} & \textbf{PDF} & \textbf{Knows} & \textbf{$\Delta$} & \textbf{LLM} & \textbf{LLM} & \textbf{Tok} & \textbf{Lat.} \\
& & & & & \textbf{PDF} & \textbf{Knows} & $\downarrow$ & $\downarrow$ \\
\midrule
Qwen-0.8B & W & 19 & \textbf{47} & +29 & 24 & \textbf{75} & 29\% & 4.6$\times$ \\
Qwen-2B & W & 25 & \textbf{67} & +42 & 25 & \textbf{77} & 29\% & 3.4$\times$ \\
MiMO-Flash & M & \textbf{76} & 66 & $-$10 & 83 & 78 & 69\% & 1.2$\times$ \\
Qwen-27B & M & 54 & 54 & 0 & 50 & 46 & 63\% & 1.0$\times$ \\
MiMO-Pro & S & \textbf{85} & 71 & $-$14 & 78 & 73 & 86\% & 1.3$\times$ \\
Kimi & S & 51 & \textbf{61} & +10 & 66 & 68 & 70\% & 0.9$\times$ \\
\bottomrule
\end{tabular}%
}
\end{table}

The LLM-judge column reveals that keyword scoring underestimates weak-model Knows accuracy by up to 32 pp (0.8B: 47\%$\rightarrow$75\%). Error attribution on 160 wrong Knows answers categorizes failures into seven types: partially correct (23.8\%), scoring false negatives (18.8\%), genuinely wrong (16.2\%), missing numeric values (15.6\%), truncated answers (12.5\%), different but valid interpretation (8.8\%), and sidecar coverage gaps (3.8\%). Combined, 42.6\% of ``errors'' are not actual errors. Zero hallucination was observed; weak models on PDF produce garbled continuations, while their Knows answers are structurally sound but miss keyword patterns.

The Knows+Fallback condition (tested on 15 original papers) achieves the highest accuracy for five of six models: Qwen-0.8B 66\%, Qwen-2B 77\%, Flash 79\%, Qwen-27B 84\%, Pro 73\%, Kimi 75\%. Fallback was triggered for 10--31 of 100 questions depending on the model (Kimi triggered most frequently at 31, Pro least at 10), indicating that sidecar coverage is sufficient for 69--90\% of questions without PDF access.

Latency speedup is substantial for local weak models (3.4--4.6$\times$ due to reduced context) but marginal for API-served models (0.9--1.3$\times$) where network overhead dominates.

\begin{center}
\fbox{\begin{minipage}{0.9\columnwidth}
\textbf{Takeaway (weak $\approx$ medium).} LLM-judge scoring confirms weak models reading sidecars (75--77\%) approach stronger models reading PDFs (78--83\%). Fallback mode recovers accuracy gaps for five of six models.
\end{minipage}}
\end{center}

\subsubsection{E2/E3: Token Efficiency and Latency}

Token reduction ranges from 29\% (weak models, whose tokenizer truncates long PDFs at 8K context) to 86\% (MiMO-V2-Pro: 2,856K$\rightarrow$401K tokens), with MiMO-V2-Flash at 69\% (1,407K$\rightarrow$441K). Latency scales proportionally for local models: Qwen-0.8B drops from 10.3s to 2.2s per question (4.6$\times$), Qwen-2B from 6.7s to 2.0s (3.4$\times$). API models show minimal speedup (1.0--1.3$\times$) because network latency dominates compute time. Even where accuracy is comparable, the 29--86\% token reduction translates directly to proportional cost savings for API-served agents processing papers at scale.

\subsubsection{E4: Review Traceability}

We measure per-weakness ID references in agent-generated reviews across 15 papers and four models (Table~\ref{tab:e4}).

\begin{table}[t]
\centering
\caption{E4 review traceability (15 papers, 4 models). Trace = per-weakness traceability; WkIDs = avg ID refs in weakness section; Comp = completeness on 3 papers with ground-truth weaknesses.}
\label{tab:e4}
\resizebox{\columnwidth}{!}{%
\begin{tabular}{llrrrrr}
\toprule
\textbf{Model} & \textbf{Cond.} & \textbf{Comp$\uparrow$} & \textbf{Trace$\uparrow$} & \textbf{WkIDs$\uparrow$} & \textbf{Struct$\uparrow$} & \textbf{Tok$\downarrow$} \\
\midrule
\multirow{2}{*}{MiMO-Pro} & PDF & 67\% & 0\% & 0.0 & 98\% & 160K \\
& Knows & 67\% & \textbf{91\%} & \textbf{3.7} & 100\% & 91K \\
\midrule
\multirow{2}{*}{MiMO-Flash} & PDF & 80\% & 0\% & 0.0 & 100\% & 156K \\
& Knows & 67\% & \textbf{73\%} & \textbf{3.2} & 100\% & 84K \\
\midrule
\multirow{2}{*}{Qwen-27B} & PDF & 60\% & 0\% & 0.0 & 77\% & 193K \\
& Knows & 53\% & \textbf{64\%} & \textbf{2.6} & 73\% & 117K \\
\midrule
\multirow{2}{*}{Kimi-K2.5} & PDF & 67\% & 0\% & 0.0 & 62\% & 163K \\
& Knows & 53\% & \textbf{80\%} & \textbf{4.3} & 68\% & 95K \\
\bottomrule
\end{tabular}%
}
\end{table}

\begin{center}
\fbox{\begin{minipage}{0.9\columnwidth}
\textbf{Takeaway (traceability).} PDF reviews contain zero per-weakness ID references across all four models. Knows reviews achieve 64--91\% traceability with 2.6--4.3 IDs per weakness section. Completeness (measured on 3 papers with ground-truth weaknesses) is comparable across conditions (53--80\%), indicating the sidecar does not harm weakness identification.
\end{minipage}}
\end{center}

\smallskip\noindent\textbf{Disclosure (prompt parity).} The PDF condition was prompted to generate a free-form review, whereas the Knows condition was prompted to cite sidecar statement IDs. Part of the 0\% to 64--91\% traceability gap is therefore attributable to a prompt-design asymmetry rather than to the metadata format alone. The reported numbers should be read as an upper bound on the effect. The matched-output consumption prompt that removes this asymmetry is validated in \S\ref{sec:e10_matched} (E10); a full numerical rerun under the matched protocol is scoped out of v0.9 and is scheduled as post-v0.9 work.

\subsubsection{E5: Consistency Validation}

We inject three types of corruption into 15 sidecars and test whether \texttt{knows-lint} detects them (Table~\ref{tab:e5}).

\begin{table}[t]
\centering
\caption{E5 lint-based detection rates on 15 papers. Structural corruption is caught; semantic corruption is not.}
\label{tab:e5}
\resizebox{\columnwidth}{!}{%
\begin{tabular}{lccl}
\toprule
\textbf{Injection} & \textbf{Applicable} & \textbf{Detected} & \textbf{Mechanism} \\
\midrule
Omit limitation & 15/15 & \textbf{15/15} & Broken cross-refs \\
Wrong number & 10/15 & 0/10 & Valid value \\
Inflate confidence & 8/15 & 0/8 & Valid enum \\
\bottomrule
\end{tabular}%
}
\end{table}

\begin{center}
\fbox{\begin{minipage}{0.9\columnwidth}
\textbf{Takeaway (lint boundary).} Deterministic lint catches 100\% of structural corruption but 0\% of semantic corruption. This clean separation motivates LLM-based semantic verification as future work. A real DP-SGD epsilon-accuracy inversion error discovered during E1 confirms the semantic gap.
\end{minipage}}
\end{center}

\subsubsection{E6: Cross-Record Traceability}

We test cross-record reasoning with 60 questions spanning paper--review sidecar pairs (15 papers, 4 models, 4 questions each; Table~\ref{tab:e6}).

\begin{table}[t]
\centering
\caption{E6 cross-record traceability (480 total results). Ratio = Knows IDs / PDF IDs per question. This experiment measures paper-to-review linkage via \texttt{record\_id\#local\_id} references; it does not evaluate unstructured cross-paper reasoning.}
\label{tab:e6}
\resizebox{\columnwidth}{!}{%
\begin{tabular}{lrrrrr}
\toprule
\textbf{Model} & \textbf{PDF Acc} & \textbf{Knows Acc} & \textbf{PDF IDs} & \textbf{Knows IDs} & \textbf{Ratio} \\
\midrule
MiMO-Pro & 98\% & 98\% & 1.3 & \textbf{5.6} & 4.3$\times$ \\
Flash & 95\% & 97\% & 1.2 & \textbf{5.1} & 4.3$\times$ \\
Qwen-27B & 92\% & 93\% & 1.6 & \textbf{8.8} & 5.5$\times$ \\
Kimi & 52\% & 67\% & 0.2 & \textbf{3.5} & 17.5$\times$ \\
\bottomrule
\end{tabular}%
}
\end{table}

\begin{center}
\fbox{\begin{minipage}{0.9\columnwidth}
\textbf{Takeaway (cross-paper).} Accuracy is comparable for strong models, but Knows answers contain 4--17$\times$ more ID references, and the IDs make multi-document reasoning chains traceable.
\end{minipage}}
\end{center}

\subsubsection{E7: LLM Sidecar Generation Quality}

We evaluate whether LLMs can generate valid sidecars from paper text across seven models spanning four providers (Table~\ref{tab:e7}).

\begin{table}[t]
\centering
\caption{E7 sidecar generation quality. Rel/S = relations per statement. Only Claude-family models pass lint.}
\label{tab:e7}
\resizebox{\columnwidth}{!}{%
\begin{tabular}{lccrrrrl}
\toprule
\textbf{Generator} & \textbf{YAML$\uparrow$} & \textbf{Lint$\uparrow$} & \textbf{Stmt} & \textbf{Evid} & \textbf{Rel} & \textbf{Rel/S} & \textbf{\$/sc} \\
\midrule
Opus 4.6 (ref.) & 5/5 & \textbf{5/5} & 7.0 & 4.2 & \textbf{15.6} & \textbf{2.2} & \$0.15 \\
Haiku 4.5 & 5/5 & \textbf{5/5} & 8.0 & 4.4 & 15.0 & 1.9 & \textbf{\$0.01} \\
Sonnet 4.6 & 5/5 & \textbf{5/5} & 10.2 & 4.6 & 15.0 & 1.5 & \$0.05 \\
\midrule
Kimi-K2.5 & 5/5 & 1/5 & 6.6 & 4.0 & 7.2 & 1.1 & \$0.03 \\
MiMO-Flash & 5/5 & 1/5 & 7.6 & 3.0 & 4.8 & 0.7 & \$0.001 \\
MiMO-Pro & 4/5 & 0/5 & 5.2 & 3.6 & 5.8 & 0.9 & \$0.02 \\
Qwen-27B & 4/5 & 0/5 & 3.8 & 3.4 & 4.8 & 1.0 & \$0.01 \\
\bottomrule
\end{tabular}%
}
\end{table}

\begin{center}
\fbox{\begin{minipage}{0.9\columnwidth}
\textbf{Takeaway (generation gap).} All three Claude-family models achieve 100\% lint pass; no non-Claude model exceeds 20\%.
\end{minipage}}
\end{center}

We further validate sidecar quality through a \emph{generator--consumer cross-evaluation} (reported in the two paragraphs below), measuring answer accuracy when an agent reads sidecars produced by different generators. The setup inverts the structural-validity lens above and asks instead whether structurally valid sidecars from different generators convey enough content to support accurate downstream answers.

\smallskip\noindent\textbf{One-shot generation.}
Four non-Claude models generate sidecars via a single API call with no lint feedback. Even when consumed by a strong model (Opus), these sidecars achieve only 23--50\% accuracy due to hallucinated values, missing claims, and shallow coverage. Shannon (formal mathematics) scores 0\% across all four generators.

\smallskip\noindent\textbf{Agent-assisted generation.}
Three Claude models generate sidecars with lint-feedback iteration via Claude Code CLI. A 5-paper subset initially suggested Haiku matches Opus (100\% vs 100\%), but a full 20-paper validation reveals a significant gap: Opus 88.6\%, Haiku dense 72.9\% ($-$15.7\,pp), Haiku non-dense 64.3\% ($-$24.3\,pp). Haiku performs well on short papers (Einstein 80\%=80\%, Watson-Crick 100\%=100\%) but fails catastrophically on complex papers (Terzaghi 0\% vs Opus 80\%). Dense mode improves Haiku by +8.6\,pp but does not close the gap. Sonnet evaluation is pending.

\begin{center}
\fbox{\begin{minipage}{0.9\columnwidth}
\textbf{Takeaway (cross-evaluation).} Agent-assisted generation ensures structural validity (lint pass) but does not guarantee content completeness. Cheaper models extract less substance from complex papers. For large-scale generation, Opus remains the quality leader; Haiku is viable only for short or structurally simple papers.
\end{minipage}}
\end{center}

\subsubsection{E8: Coverage--Accuracy--Token Ablation}

We ablate sidecar components to map the accuracy--token Pareto frontier (15 papers, 4 models, 5 conditions, 2,119 results; Table~\ref{tab:e8}).

\begin{table}[t]
\centering
\caption{E8 ablation (2,119 results). Eff. = accuracy per 1K input tokens.}
\label{tab:e8}
\resizebox{\columnwidth}{!}{%
\begin{tabular}{lrrrc}
\toprule
\textbf{Condition} & \textbf{Acc$\uparrow$} & \textbf{Avg Tok$\downarrow$} & \textbf{Saved$\uparrow$} & \textbf{Eff.} \\
\midrule
PDF & 59\% & 9,955 & --- & 5.9 \\
Full sidecar & \textbf{59\%} & 4,463 & 55\% & 13.2 \\
$-$relations & 58\% & 3,998 & 60\% & 14.5 \\
$-$evidence & 57\% & 3,057 & 69\% & 18.6 \\
Stmts-only & 52\% & 693 & \textbf{93\%} & \textbf{75.0} \\
\bottomrule
\end{tabular}%
}
\end{table}

Dropping relations costs just 1 pp: relations aid traceability (E4, E6), not comprehension. Dropping evidence costs 2 pp, reflecting the value of numeric observations. Statements-only retains 88\% of full-sidecar accuracy at 7\% of PDF tokens. The aggregate hides model-level divergence: Kimi-K2.5 rises 27\% (PDF) $\rightarrow$ 38\% (sidecar), while MiMO-V2-Pro (73\%$\rightarrow$69\%), Flash (71\%$\rightarrow$69\%), and Qwen-27B (67\%$\rightarrow$61\%) slightly prefer PDF.

\begin{center}
\fbox{\begin{minipage}{0.9\columnwidth}
\textbf{Takeaway (Pareto frontier).} Full sidecar matches PDF accuracy (59\%) with 55\% fewer tokens. Statements-only retains 88\% of accuracy with 93\% fewer tokens, achieving 12.7$\times$ higher token efficiency than PDF.
\end{minipage}}
\end{center}

\subsubsection{E9: Statement Granularity Ablation}

The preceding experiments use sidecars with uniform granularity ($\sim$7 statements per paper). To test whether poor performance on certain disciplines reflects YAML representation limits or insufficient coverage, we created \emph{dense} sidecar variants (15--25 statements) for eight underperforming papers, targeting questions previously marked \texttt{sidecar\_answerable: false}.

\begin{table}[t]
\centering
\caption{E9 statement granularity ablation. Accuracy (\%) for original ($\sim$7 stmts) vs dense (15--25 stmts) sidecars across four model tiers. $\Delta$ = dense $-$ original.}
\label{tab:e9}
\resizebox{\columnwidth}{!}{%
\begin{tabular}{lrrrrrrrr}
\toprule
& \multicolumn{2}{c}{\textbf{0.8B}} & \multicolumn{2}{c}{\textbf{2B}} & \multicolumn{2}{c}{\textbf{Medium}} & \multicolumn{2}{c}{\textbf{Strong}} \\
\cmidrule(lr){2-3}\cmidrule(lr){4-5}\cmidrule(lr){6-7}\cmidrule(lr){8-9}
\textbf{Paper} & Orig & Dense & Orig & Dense & Orig & Dense & Orig & Dense \\
\midrule
G\"{o}del & 40 & \textbf{100} & 60 & \textbf{100} & 60 & \textbf{100} & 60 & \textbf{100} \\
Shannon & 40 & \textbf{60} & 20 & \textbf{40} & 20 & \textbf{80} & 20 & \textbf{60} \\
Kahneman & 38 & 38 & 75 & 50 & 63 & \textbf{88} & 38 & \textbf{75} \\
Semmelweis & 20 & 0 & 20 & \textbf{60} & 80 & \textbf{100} & 60 & \textbf{80} \\
Gettier & 63 & 63 & 63 & \textbf{75} & 63 & 63 & 63 & \textbf{75} \\
\bottomrule
\end{tabular}%
}
\end{table}

Dense sidecars improve accuracy for medium and strong models on most papers (+13 to +60 pp where improvement occurs), with G\"{o}del reaching 100\% across all four tiers. For weak models, dense sidecars help where the additional statements cover previously missing content (G\"{o}del: +60 pp for 0.8B) but show no benefit when the paper's reasoning structure exceeds model capacity (Kahneman: 0 pp for 0.8B). Comparing token consumption across all three representations: non-dense sidecars save 77\% of tokens versus PDF (avg.\ 4.7K vs 19.9K), dense sidecars save 51\% (avg.\ 9.7K vs 19.9K), while delivering +25\,pp accuracy over non-dense for medium models. Short papers ($<$2K words) are an exception where schema overhead makes the sidecar larger than the PDF itself.

\begin{center}
\fbox{\begin{minipage}{0.9\columnwidth}
\textbf{Takeaway (Granularity).} Sidecar effectiveness is driven primarily by statement granularity, not YAML format limitations. Dense sidecars eliminate discipline-specific performance gaps for medium and strong models. The uniform $\sim$7-statement template used in E1--E8 was a generation artifact, not a specification constraint.
\end{minipage}}
\end{center}

We additionally test whether dense sidecars benefit from PDF fallback (E9b). Across 16 paper--model combinations (8 papers $\times$ 2 models), fallback improves accuracy in only 3 cases (Kahneman/Flash +12.5\,pp, Shannon/Flash +20\,pp, Semmelweis/Kimi +20\,pp) while \emph{hurting} in 2 cases where hedging detection triggers incorrectly on verbose but correct answers. Dense sidecars are sufficiently self-contained that the fallback mechanism adds latency without reliable benefit.

\subsubsection{E10: Matched-Output Protocol Validation}
\label{sec:e10_matched}

To separate the metadata-format effect from prompt-design asymmetries noted in E4, we introduce the matched-output consumption prompt, in which both the Knows and the PDF conditions emit a single JSON object with per-claim verbatim quotes and page numbers. The v0.9 release ships this experiment at pilot scale: 10 questions across three strong consumers (MiMO-V2-Flash, Claude Opus, Kimi-K2.5) in both conditions, for a total of 60 cells.

\begin{table}[h]
\centering
\caption{E10 pilot gate check. Thresholds (parse-rate $\geq$95\%, page-echo $\geq$90\%, quote-verbatim $\geq$80\%) are locked in the pre-registered release-rules document before the run.}
\label{tab:e10_matched}
\resizebox{\linewidth}{!}{%
\begin{tabular}{lcccc}
\toprule
\textbf{Consumer} & \textbf{OK/20} & \textbf{Parse} & \textbf{Page-echo} & \textbf{Quote-verbatim} \\
\midrule
Kimi-K2.5        & 20/20 & 100\% & 100\% & 100\% \\
MiMO-V2-Flash    & 20/20 & 100\% & 100\% & 95\% \\
Claude Opus      & 20/20 & 100\% & 100\% & 95\% \\
\midrule
Pooled           & 60/60 & 100\% & 100\% & 97\% \\
\bottomrule
\end{tabular}
}
\end{table}

All three consumers pass every gate. The 5\% non-verbatim cases on Flash and Opus are minor punctuation or formula-rendering variants that match under Unicode normalization. The status distribution for Opus (14 \texttt{answer}, 3 \texttt{abstain}, 2 \texttt{not\_found}, 1 \texttt{partial}) demonstrates the safety property of Algorithm~\ref{alg:consume}: when the sidecar does not support the question, the consumer honestly refuses rather than fabricating an answer, and Knows+Fallback then defers to the PDF (Appendix~\ref{app:examples}, Example 2). The pilot validates that the matched-output protocol is reliably executable on heterogeneous consumer stacks. A full 840-call rerun (140 questions $\times$ 3 consumers $\times$ 2 conditions) to produce numerical accuracy deltas versus E1/E4 is explicit future work and is scoped out of the v0.9 release; its release rules and outcome--action decision tree are pre-registered in the release-rules document distributed alongside this report.

\subsection{Discussion and Analysis}
\label{sec:discussion}

\subsubsection{Structural vs.\ Semantic Verification}

E5 reveals a clean capability boundary. Deterministic lint catches all structural corruption because entity removal breaks cross-reference integrity against the schema graph. Semantic corruption (wrong numbers, inflated confidence) yields schema-valid records and needs future LLM-based verification; the DP-SGD epsilon-accuracy inversion in E1, where a hand-curated sidecar mis-mapped a value, confirms this in practice. Structured malice is thus more dangerous than ambiguous prose---a schema-valid sidecar with fabricated numbers could automate misinformation at scale---motivating the two-layer verification design.

\subsubsection{When Does Knows Help?}

The interaction between paper length and model capacity determines when sidecar-based retrieval outperforms direct PDF reading. For weak models (0.8B--2B), the sidecar provides a structured, token-efficient representation that compensates for limited context windows at all paper lengths. Weak models produce garbled continuations on long PDFs (not wrong answers, but degenerate output), making the PDF baseline a test of context-length capability rather than comprehension. For medium and strong models, the sidecar matches accuracy on short-to-medium papers while saving 35--86\% tokens, but loses fine-grained details present only in the full text of longer papers. The fallback mode recovers this gap: five of six models achieve their highest accuracy in the Knows+Fallback condition.

Question difficulty stratification across five tiers reveals distinct patterns. Factual-detailed questions (43 of 140, requiring multi-sentence factual answers) represent the sweet spot where all model tiers benefit (weak +30 pp, medium +9 pp, strong +21 pp). Factual-numeric questions (14 of 140) show the largest weak-model gains (+46 pp) but no benefit for strong models that locate numbers directly in PDFs. Reasoning questions (23 of 140) show consistent gains across all tiers (weak +39 pp, medium +7 pp, strong +7 pp), suggesting that the sidecar's pre-extracted claim-evidence structure reduces inference burden even for capable models.

\subsubsection{Limitations}

All sidecars and benchmark questions were authored by the same LLM (Claude Opus), creating circular evaluation bias. The E1--E8 sidecars also use a uniform $\sim$7-statement template regardless of paper complexity, which E9 shows underestimates optimal granularity for complex papers by up to 57\,pp. Numeric questions mitigate this concern because their ground truth is objectively verifiable from the source text. The keyword scoring method underestimates accuracy by up to 32 percentage points for weak models, as confirmed by the LLM-judge experiment; we report both scores throughout. All 20 papers are well-known classics; modern papers with supplementary materials, code repositories, and multi-part structures remain untested. The evaluation covers six models from three providers; five additional models were tested (Qwen3.5-9B, Gemma-3-1B, Qwen3-1.7B, LFM-2.5-1.2B, LFM-2.5-1.2B-Think) but excluded due to insufficient paper coverage for fair comparison. Other architectures (Llama, Gemini, GPT) are not tested. Only English-language papers are included (four originally in German, evaluated in English translation). E7 uses only the first 15K characters per paper due to API context limits.

\section{Case Studies and Applications}

This section illustrates how \method{} can be used in practice beyond single-paper question answering. We focus on three representative settings: large-scale literature intelligence, human-agent collaborative knowledge work, and multi-agent research ecosystems.

\subsection{Case 1: Large-Scale Literature Intelligence}

The first application is large-scale literature intelligence: conference analysis, yearly trend tracking, survey bootstrapping, and emerging-topic detection across thousands of papers. In these settings, title- and abstract-level metadata are often too coarse to support method-level or evidence-aware synthesis, while full PDF or HTML ingestion is too costly to run continuously at scale. \method{} provides an intermediate representation that is substantially richer than bibliographic metadata yet substantially cheaper to consume than full text.

In practice, this enables workloads such as conference-scale literature analysis. For example, the current \method{} deployment has been used to generate ICLR 2026 insights over more than five thousand papers\footnote{\url{https://knows.academy/insights\#iclr26/topics/en}}, including trend-oriented views that go beyond abstract-level topic summaries. The same pattern extends to field-level survey drafting, conference summaries, trend reports, and longitudinal analyses of how methods, tasks, and evaluation practices evolve over time.

\subsection{Case 2: Human-Agent Collaborative Knowledge Work}

Human-agent collaborative knowledge work goes beyond isolated question answering: the same paper must be adapted to different downstream tasks. A researcher drafting related work wants concise comparisons of methods, assumptions, and limitations; one designing experiments prefers evidence-centered views that foreground results, settings, and caveats; a knowledge-graph builder wants entities and typed relations. Reader-oriented documents force humans and agents to re-derive these task-specific views from full text each time. \method{} exposes a task-addressable intermediate representation, letting the same paper be reorganized by user intent rather than document order.

\subsection{Case 3: Agent-Native Research Ecosystems}

The third application is agent-native research ecosystems, in which papers are no longer consumed only by human readers or single assistants, but also by research agents that participate in longer workflows. These workflows may include literature monitoring, review analysis, hypothesis generation, experiment planning, or coordination among multiple agents operating in shared environments. As agent-led and agent-assisted research initiatives become more common, papers need to function not only as documents to read but also as structured research objects that can be passed between tools, agents, and memory systems.

\method{} makes this literature interaction modular and auditable, extending beyond single-agent paper question answering to emerging pipelines that combine scholarly platforms such as OpenReview\footnote{\url{https://openreview.net}} with agent-native discussion spaces such as Moltbook\footnote{\url{https://www.moltbook.com}} and shared memory systems. There, a structured sidecar is the substrate that connects document ingestion, grounded discussion, and long-horizon research memory. As research workflows become more agent-mediated, papers need to be exchangeable not only as documents to read but also as structured objects that agents can monitor, discuss, compare, and accumulate.

\section{Agent-Native Publishing: Vision and Trajectory}
\label{sec:vision}

The evaluation and the deployment evidence in this report target an immediate, tractable goal: make today's PDF-centric literature cheaper and more reliable for agents to consume. The longer trajectory, however, is an inversion of the information flow that currently governs scientific communication.

\begin{figure}[t]
  \centering
  \includegraphics[width=\linewidth]{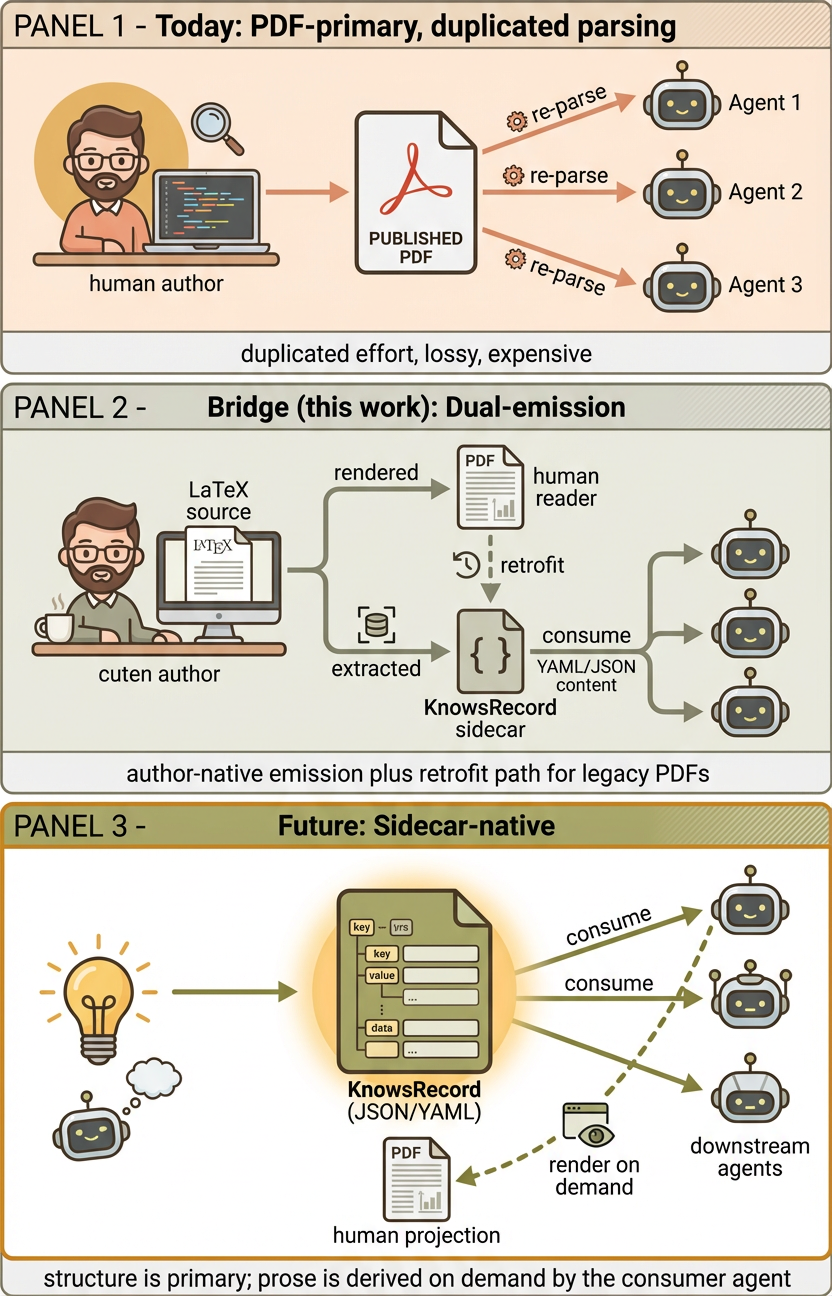}
  \caption{Three phases of the agent-native publishing trajectory. \emph{Today}: authors emit PDFs; every downstream agent independently re-parses the same unstructured prose. \emph{Bridge} (this work): a companion sidecar translates the PDF into structured form, which agents consume directly while humans keep the PDF. \emph{Future}: authors (or autonomous research agents) emit an agent-native record first; human-readable PDFs are rendered on demand as a projection of that record. The information flow is reversed: structure is primary, prose is derived.}
  \label{fig:vision}
\end{figure}

\smallskip\noindent\textbf{Today: dual-parsing is the bottleneck.}
Authors write papers in a format optimized for human eyes, yet an expanding fraction of downstream readers are LLM-based agents. Every agent reconstructs structure from the same prose (extracting claims, resolving citations, mapping evidence to assertions) with no shared intermediate. The cost is duplicated compute, lossy extractions, and mutually incompatible reconstructions. The experiments in this report quantify this cost: weak models burn 29--86\% more tokens for equal or worse comprehension than sidecar-augmented counterparts, and traceability collapses to 0\% under free-form PDF prompts.

\smallskip\noindent\textbf{Bridge: dual-emission from the same source.}
A near-term step, which this report already enables, is to emit both forms from a single source. Authoring toolchains already produce LaTeX, Word, or Markdown as an intermediate; our \texttt{knows-gen} scaffolder shows that a KnowsRecord can be synthesized from such a source at authoring time. The PDF remains the human-facing artifact; the sidecar becomes the agent-facing artifact. Publishers, preprint servers, and paper-authoring tools can begin shipping sidecars immediately, without altering the human reading experience.

\smallskip\noindent\textbf{Future: sidecar-native publishing.}
The autonomous research agents catalogued in \S\ref{sec:autonomous-research} already author and consume research in structured internal representations, and PDF output is generated from those representations rather than the other way around. The trajectory is no longer speculative at the institutional level either: Agents4Science~\cite{agents4science2025} treats AI agents as the primary authors, reviewers, and presenters of a peer-reviewed research venue, which makes the question of a shared agent-native publishing format an immediate one rather than a hypothetical one. Extrapolating from that shift, the endpoint is a sidecar-native publishing pipeline in which idea $\rightarrow$ experiment $\rightarrow$ typed record is the primary authoring path, and a human-readable rendering (PDF, web page, narrated audio) is produced on demand by a translator agent when a human reader actually needs one. Review, in the same frame, is a cross-record relation: one agent-produced critique linked by typed edges to another agent-produced paper, with prose rendered only when a human reviewer or author engages. Humans read rendered artefacts; the long-run scholarly graph is built directly from the structured records.

\smallskip\noindent\textbf{What this implies for adoption.}
The three phases are cumulative rather than alternatives. A sidecar authored for the Bridge phase carries unchanged into the Future phase, so early deployment produces artefacts that remain useful even as the publishing model shifts. The Bridge phase is already defensible on its own: the evaluation in this report and the indexed corpus at \url{https://knows.academy/} show measurable token savings and higher traceability on tens of thousands of papers today. Waiting for the Future phase to arrive in a finished form is not a neutral choice; it is a choice to accumulate no structured scholarly record until then.

\section{Conclusion}

We presented \method{}, a companion sidecar specification for research artifacts that provides agents with structured, schema-validated metadata without modifying the original publication. Evaluation across 20 papers, 14 disciplines, and six LLMs reports three primary findings: (1) weak models gain +29 to +42 percentage points accuracy from sidecar-assisted retrieval, approaching medium-model PDF performance; (2) full sidecars attain accuracy comparable to PDF-reading on strong consumers with 55\% fewer tokens, while minimal statement-only sidecars retain 88\% of full-sidecar accuracy at 93\% token savings; and (3) per-weakness ID references rise from 0\% under a free-form PDF prompt to 64--91\% under a Knows citation-requesting prompt, with the portion attributable to prompt asymmetry vs metadata format under re-measurement (Disclosure, \S\ref{sec:experiments}).

\smallskip\noindent\textbf{Release status and roadmap.}
The v0.9 specification frozen in this report covers the core schema (30 root-level fields, 23 entity definitions), the normative consumption protocol (Algorithm~\ref{alg:consume}), the \texttt{paper@1} and \texttt{review@1} profiles, the reference linter and generator, the canonical consumption prompt, and the matched-output protocol validated in \S\ref{sec:e10_matched}. 
v1.0 is gated on three in-progress studies whose protocols are locked here: the full 840-call matched-output rerun (E10), the audit-trace validity rubric, and the coverage-conditioned analysis; their paper-level consequences are pre-registered in the accompanying release-rules document so v1.0 ships with calibrated rather than post-hoc claims. Post-v1.0 directions include \texttt{dataset@1}, \texttt{model@1}, and \texttt{benchmark@1} profiles; a cross-record \texttt{replaces}/\texttt{retracts} lifecycle test suite; LLM-based semantic verification to complement deterministic lint; an adversarial semantic-corruption probe; a community-curated Sidecar Hub; a calibrated cross-model confidence probe replacing the hedging-phrase fallback with the $\tau$ threshold of Algorithm~\ref{alg:consume}; and live-cycle evaluation of author-native workflows in conference review pipelines.

\bibliographystyle{IEEEtran}
\bibliography{refs}

\appendices

\section{Schema Reference Summary}
\label{app:schema}

The complete JSON Schema v0.9 is released alongside the specification at \url{https://knows.academy/}; this appendix summarizes the root-level structure and provides a minimal valid record. Unless stated otherwise, every root object enforces \texttt{additionalProperties: false} so that unknown fields cause schema validation to fail; forward-compatible extension is only allowed through the \texttt{x\_extensions} object.

\begin{table}[h]
\centering
\resizebox{\linewidth}{!}{%
\begin{tabular}{@{}lll@{}}
\toprule
\textbf{Field} & \textbf{Type} & \textbf{Purpose} \\
\midrule
\texttt{\$schema} & string & Fixed JSON Schema URI \\
\texttt{knows\_version} & semver & Target spec version \\
\texttt{record\_id} & string & Globally unique sidecar ID \\
\texttt{profile} & string & Domain namespace (\texttt{paper@1}, \texttt{review@1}, \ldots) \\
\texttt{subject\_ref} & string & Local ID of the primary artifact \\
\texttt{title} & string & Sidecar title \\
\texttt{summary} & string & One-paragraph overview \\
\texttt{authors} & array & Structured author records \\
\texttt{license} & SPDX & Licence of the sidecar record \\
\texttt{coverage} & object & Statement/evidence completeness \\
\texttt{artifacts} & array & Papers, repos, datasets, cited works \\
\texttt{statements} & array & Claims, assumptions, limitations, \ldots \\
\texttt{evidence} & array & Table cells, figures, experiment runs \\
\texttt{relations} & array & Typed edges across entities \\
\texttt{provenance} & object & Authoring origin, actor, method \\
\texttt{version} & object & Spec, record, and source versions \\
\texttt{freshness} & object & \texttt{as\_of}, update policy, staleness \\
\bottomrule
\end{tabular}
}
\caption{Root-level required fields of a KnowsRecord v0.9. Optional fields include \texttt{abstract}, \texttt{venue}, \texttt{year}, \texttt{keywords}, \texttt{record\_status}, \texttt{replaces}, \texttt{actions}, \texttt{resources}, and \texttt{extensions}.}
\label{tab:root-fields}
\end{table}

\noindent A minimal valid record contains the 17 required fields above and satisfies the cross-reference rule that \texttt{subject\_ref} names an entry of the \texttt{artifacts} array (the \texttt{\$schema} URI, \texttt{https://knows.dev/schema/record-0.9.json}, is prepended by every conforming writer and is omitted from the listing below for brevity):

\begin{lstlisting}[style=compact]
knows_version: 0.9.0
record_id: 10.1234/example
profile: paper@1
subject_ref: art:paper
title: Example Sidecar
summary: Minimal record for illustration.
authors:
  - name: Anonymous
    affiliation: Independent
license: CC-BY-4.0
coverage: {statements: partial, evidence: partial}
artifacts:
  - id: art:paper
    artifact_type: paper
    role: subject
    title: Example Paper
    identifiers: {doi: '10.1234/example'}
    representations:
      - id: rep:pdf
        media_type: application/pdf
        locator: {type: path, value: 'paper.pdf'}
statements: []
evidence: []
relations: []
provenance:
  origin: author
  actor: {name: Example Author, type: person}
  generated_at: '2026-04-18T00:00:00Z'
version: {spec: 0.9.0, record: '1', source: 'v1'}
freshness:
  as_of: '2026-04-18T00:00:00Z'
  update_policy: versioned
\end{lstlisting}

Table~\ref{tab:root-fields} lists only the root level. Each of \texttt{artifacts}, \texttt{statements}, \texttt{evidence}, \texttt{relations}, and \texttt{actions} carries its own entity schema with additional required fields (e.g., every \texttt{Statement} requires \texttt{id}, \texttt{statement\_type}, \texttt{text}, \texttt{about\_ref}, \texttt{status}). Complete field-by-field documentation is in the repository.

\section{Quickstart Walkthrough}
\label{app:quickstart}

A KnowsRecord can be generated, validated, and consumed in three commands. The examples below use the reference implementation \texttt{knows-sidecar}; the same operations are expressible in any language because both the schema and the linter rules are fully specified.

\smallskip\noindent\textbf{Installation.}

\begin{lstlisting}
pip install knows-sidecar
\end{lstlisting}

\smallskip\noindent\textbf{Step 1: Generate.}
Two paths are supported. \emph{Author-native} generation scaffolds from the author's manuscript source (LaTeX, Markdown, or any supported format) and emits a KnowsRecord with template placeholders the author completes (typical effort: 15 minutes per conference paper). \emph{Retrofit} generation extracts structured metadata from an existing PDF via an LLM, producing a record that is statistically consistent with the source but still requires a lint pass. The two paths emit the same schema and are interchangeable for downstream consumers.

\begin{lstlisting}
# author-native (from LaTeX)
knows gen --from-latex paper.tex \
          --out paper.knows.yaml

# retrofit (from PDF, LLM-assisted)
knows gen --from-pdf paper.pdf \
          --model haiku \
          --out paper.knows.yaml
\end{lstlisting}

\smallskip\noindent\textbf{Step 2: Validate.}
The linter performs the seven deterministic checks described in \S4 and reports each violation with a stable error code, enabling CI integration.

\begin{lstlisting}
knows lint paper.knows.yaml
# => PASS (7/7 checks)
\end{lstlisting}

\smallskip\noindent\textbf{Step 3: Consume.}
An agent implements the protocol of Algorithm~\ref{alg:consume}. Given the composition \texttt{Compose($R, \mathcal{A}$)} and a query $q$, the minimal consumer retrieves top-$k$ statements and evidence, resolves their anchors, and returns an answer with a citation trace:

\begin{lstlisting}
from knows_sidecar import load, Consumer

record = load("paper.knows.yaml")
agent = Consumer(record, mode="fallback", tau=0.7)
answer, trace = agent.ask(
  "What accuracy did the method achieve "
  "on the main benchmark?"
)
\end{lstlisting}

\noindent The returned \texttt{trace} enumerates the statements and evidence entries used to compose the answer, together with the \texttt{source\_anchors} resolved during retrieval. Consumers that operate in \texttt{Knows-only} mode emit the trace without ever loading the PDF; consumers in \texttt{Knows+Fallback} mode additionally include the PDF passages consulted when the sidecar-only confidence fell below $\tau$.

\smallskip\noindent\textbf{Review-as-sidecar.}
Review records (\texttt{profile: review@1}) link weakness statements to specific claims via cross-record relations of the form \texttt{record\_id\#stmt:c1}; Section~5 reports accuracy, traceability, and consistency.

\section{End-to-End Worked Examples}
\label{app:examples}

The two transcripts below are verbatim outputs from the E10 matched-output pilot run, in which a \textsc{Claude Opus} consumer executes Algorithm~\ref{alg:consume} under the canonical prompt \texttt{consume-prompt.md} v1.1. They are included to make the protocol concrete: the JSON is exactly what the agent returns, and the evidence quotes are verbatim substrings of the supplied context (sidecar or PDF).

\subsection{Example 1: \emph{Knows-only} success}

Paper: \emph{Taxonomy of Educational Objectives} (Bloom, 1956). \\
Question: ``What are the six levels of Bloom's original taxonomy in order from lowest to highest, and what cognitive activity does each level represent?''
Context supplied: the paper's KnowsRecord (a 17\,KB YAML file). The PDF is not consulted.

\begin{lstlisting}[style=compact]
{
  "item_id": "bloom:bloom-L1-1",
  "status": "answer",
  "confidence": 0.98,
  "answer": "Bloom's original taxonomy has six levels
    ordered from lowest to highest: (1) Knowledge -
    recall of specifics, universals, methods,
    processes, patterns, structures, or settings;
    (2) Comprehension - understanding meaning;
    translation, interpretation, extrapolation;
    (3) Application - use of abstractions in
    particular and concrete situations;
    (4) Analysis - breakdown into constituent
    elements; relationships, organizational
    principles; (5) Synthesis - putting elements
    together to form a new whole; unique
    communication, plan, or set of abstract
    relations; (6) Evaluation - judgments about the
    value of material and methods for given
    purposes.",
  "evidence": [
    {
      "source": "paper",
      "page": 0,
      "quote": "(1) Knowledge, (2) Comprehension,
        (3) Application, (4) Analysis,
        (5) Synthesis, and (6) Evaluation",
      "support": "direct"
    },
    {
      "source": "paper",
      "page": 0,
      "quote": "Knowledge - recall of specifics,
        universals, methods, processes, patterns,
        structures, or settings",
      "support": "direct"
    }
  ],
  "reason": "Sidecar statement stmt:c1 lists six
    ordered levels; ev:six-levels provides the
    cognitive activity description for each level
    verbatim."
}
\end{lstlisting}

The consumer returns a full taxonomy, grounded in two verbatim quotes from the sidecar (indicated by \texttt{page: 0}), with explicit provenance back to the statement and evidence identifiers. No PDF retrieval was required: Knows-only mode emits the full trace at a fraction of the full-text token cost.

\subsection{Example 2: legitimate \emph{abstain} and fallback}

Paper: \emph{Cramming more components onto integrated circuits} (Moore, 1965).
Question: ``Why does Moore argue that heat dissipation will not be a fundamental barrier to integrating tens of thousands of components on a single chip?''

\paragraph{Stage 1 (Knows-only).} The consumer receives the Moore sidecar, which captures the paper's famous scaling prediction and the economic argument but does not explicitly encode the heat-dissipation subargument. The consumer abstains:

\begin{lstlisting}
{
  "item_id": "moore:moore-L1-4",
  "status": "abstain",
  "confidence": 0.0,
  "answer": "",
  "evidence": [],
  "reason": "Sidecar does not discuss heat
    dissipation or Moore's argument that it is
    not a fundamental barrier."
}
\end{lstlisting}

This is the safety property in action: the consumer refuses to hallucinate an answer when the sidecar does not support it, preferring an honest abstention. Under Algorithm~\ref{alg:consume}, the low confidence triggers the \emph{Knows+Fallback} branch.

\paragraph{Stage 2 (fallback to PDF).} The same consumer now receives the page-tagged PDF and answers:

\begin{lstlisting}[style=compact]
{
  "item_id": "moore:moore-L1-4",
  "status": "answer",
  "confidence": 0.95,
  "answer": "Moore argues that heat dissipation
    will not be a fundamental barrier because
    integrated electronic structures are
    two-dimensional, providing a cooling surface
    close to each center of heat generation;
    additionally, as long as functions are
    confined to small areas on a wafer, the
    capacitance that must be driven is limited,
    and shrinking dimensions actually allows
    operation at higher speed for the same power
    per unit area.",
  "evidence": [
    {
      "source": "paper",
      "page": 3,
      "quote": "integrated electronic structures
        are two dimensional, they have a surface
        available for cooling close to each
        center of heat generation",
      "support": "direct"
    },
    {
      "source": "paper",
      "page": 4,
      "quote": "shrinking dimensions on an
        integrated structure makes it possible to
        operate the structure at higher speed for
        the same power per unit area",
      "support": "direct"
    }
  ],
  "reason": "Moore cites two-dimensional structure
    with nearby cooling surfaces and limited
    capacitance in small areas, enabling higher
    speed per unit power."
}
\end{lstlisting}

Pages 3 and 4 in the returned evidence correspond to \texttt{[PAGE 3]} and \texttt{[PAGE 4]} markers in the supplied context. The composite citation trace exposed by \emph{Knows+Fallback} records both the sidecar-stage abstention and the PDF-stage answer, so a downstream auditor can reconstruct exactly why the consumer fell back and which page-level evidence produced the final response.
This two-stage pattern motivates the \texttt{not\_found}/\texttt{abstain} status values in \texttt{consume-prompt.md} v1.1 and the threshold $\tau$ in Algorithm~\ref{alg:consume}: the specification asks consumers to be honest about sidecar coverage limits and defers to the PDF when necessary, rather than silently inventing answers.

\section{This Paper's Own KnowsRecord}
\label{app:dogfood}

To demonstrate the specification end-to-end, this paper is accompanied by its own KnowsRecord sidecar, produced by an LLM consumer running the Knows skill (\url{https://knows.academy/for-agents}) against the full manuscript and validated by the reference linter (zero schema errors, zero discoverability warnings). The complete record is released as \texttt{knows-record.yaml} alongside the paper; it contains 15 artifacts (the paper itself plus 14 cited works), 21 statements, 15 evidence items, 33 typed relations, and 2 executable actions. An excerpt is reproduced below, showing the primary artifact, three representative statements (a core claim, a limitation, and a method definition), two evidence items (one quantitative table result with numeric observations and one citation-backed external pointer), three relations drawn from distinct predicates, one action with an explicit safety policy, and the mandatory provenance, version, and freshness blocks.

\begin{lstlisting}[style=compact]
# schema URI: https://knows.dev/schema/record-0.9.json (omitted to keep the listing LaTeX-safe)
knows_version: 0.9.0
record_id: 10.0000/knows-v0.9-tech-report-2026
profile: paper@1
subject_ref: art:paper
title: 'Knows: Agent-Native Structured Research Representations'
authors:
- name: Guangsheng Yu
  affiliation: Independent Researcher
  role: first
- name: Xu Wang
  affiliation: Independent Researcher
  role: contributor
summary: Technical report introducing Knows, a lightweight YAML sidecar specification
  (KnowsRecord v0.9) that binds claims, evidence, provenance, and typed relations
  to existing research artifacts so LLM agents can consume them directly,
  evaluated on 140 questions across 20 papers and 14 disciplines.
license: CC-BY-4.0
coverage:
  statements: main_claims_only
  evidence: key_evidence_only
artifacts:
- id: art:paper
  artifact_type: paper
  role: subject
  title: 'Knows: Agent-Native Structured Research Representations'
  identifiers:
    url: https://knows.academy/
  representations:
  - id: rep:paper-pdf
    media_type: application/pdf
    locator:
      type: path
      value: paper.pdf
  # ... 14 more artifacts in the full record
statements:
- id: stmt:pdf-agent-bottleneck
  statement_type: claim
  modality: descriptive
  text: Reader-oriented PDF distribution creates a bottleneck for agent-assisted
    and agent-native research workflows because every LLM agent must independently
    re-extract fine-grained structure from the same prose, a process that
    is expensive, repetitive, and unstable at scale.
  about_ref: art:paper
  status: asserted
  source_anchors:
  - representation_ref: rep:paper-pdf
    locator_type: section
    locator: Section 1 Introduction
  confidence:
    claim_strength: high
    extraction_fidelity: high
  provenance:
    origin: author
    actor:
      name: Guangsheng Yu
      type: person
    generated_at: '2026-04-19T00:00:00Z'
- id: stmt:circular-evaluation-limitation
  statement_type: limitation
  modality: descriptive
  text: All sidecars and benchmark questions were authored by the same LLM
    (Claude Opus), creating a circular-evaluation bias that the numeric ground
    truth for factual-numeric questions only partially mitigates.
  about_ref: art:paper
  status: asserted
  source_anchors:
  - representation_ref: rep:paper-pdf
    locator_type: section
    locator: Section 5.3 Limitations
  confidence:
    claim_strength: high
    extraction_fidelity: high
  provenance:
    origin: author
    actor:
      name: Guangsheng Yu
      type: person
    generated_at: '2026-04-19T00:00:00Z'
- id: stmt:knows-sidecar-specification
  statement_type: method
  modality: descriptive
  text: KnowsRecord v0.9 is a YAML companion sidecar specification with 30
    root-level fields and 23 entity definitions that binds structured claims,
    evidence, provenance, and typed relations to existing research artifacts,
    validated by a deterministic schema linter and requiring no changes to
    the original publication.
  about_ref: art:paper
  status: asserted
  source_anchors:
  - representation_ref: rep:paper-pdf
    locator_type: section
    locator: Section 3 The KnowsRecord Specification
  confidence:
    claim_strength: high
    extraction_fidelity: high
  provenance:
    origin: author
    actor:
      name: Guangsheng Yu
      type: person
    generated_at: '2026-04-19T00:00:00Z'
  # ... 18 more statements in the full record
evidence:
- id: ev:e1-length-stratified-accuracy
  evidence_type: table_result
  summary: E1 accuracy stratified by paper length and model tier; weak models
    improve at every paper length (+8 pp on SHORT to +57 pp on LONG), medium
    models favour PDF on STANDARD papers (-19 pp), strong models show near-parity.
  source_anchors:
  - representation_ref: rep:paper-pdf
    locator_type: table
    locator: Table 3 (tab:e1_length)
  observations:
  - metric: weak_model_delta_short_pp
    value: 8.0
    unit: pp
  - metric: weak_model_delta_long_pp
    value: 57.0
    unit: pp
  - metric: medium_model_delta_standard_pp
    value: -19.0
    unit: pp
  provenance:
    origin: author
    actor:
      name: Guangsheng Yu
      type: person
    generated_at: '2026-04-19T00:00:00Z'
- id: ev:paper2agent-citation
  evidence_type: citation_backed
  summary: Paper2Agent is cited as a representative approach that converts
    papers into MCP tool endpoints post hoc, sharing the limitation that structure
    is reconstructed from prose at consumption time.
  source_anchors:
  - representation_ref: rep:paper-pdf
    locator_type: section
    locator: Section 2.2 LLM Agents and Paper Consumption
  observations:
  - metric: prior_finding
    qualitative_value: Paper2Agent converts papers into MCP tool endpoints
      post hoc so an agent can interact with individual papers through a retrieval
      pipeline.
  provenance:
    origin: author
    actor:
      name: Guangsheng Yu
      type: person
    generated_at: '2026-04-19T00:00:00Z'
  # ... 13 more evidence in the full record
relations:
- id: rel:method-documents-paper
  subject_ref: stmt:knows-sidecar-specification
  predicate: documents
  object_ref: art:paper
  # ... 32 more relations in the full record
actions:
- id: act:query-knows-hub
  action_type: query
  label: Query the knows.academy community sidecar hub for indexed records
  target_ref: art:knows-hub
  interface:
    kind: http
    locator: https://knows.academy/
  safety:
    sandbox_required: false
    network: true
    secrets_required: false
    side_effects: none
provenance:
  origin: author
  actor:
    name: Guangsheng Yu
    type: person
  generated_at: '2026-04-19T00:00:00Z'
  method: manual_curation
version:
  spec: 0.9.0
  record: '1'
  source: v0.9-tech-report
freshness:
  as_of: '2026-04-19T00:00:00Z'
  update_policy: versioned

\end{lstlisting}

\noindent The full 1\,084-line record accompanies this paper and passes \texttt{knows gen} $\rightarrow$ \texttt{knows lint} $\rightarrow$ \texttt{knows query} end-to-end, validating the author-native path on the paper that describes it.

\end{document}